\documentclass[conference]{IEEEtran}
\IEEEoverridecommandlockouts
\usepackage{hyperref}
\usepackage{cite}
\usepackage{makecell}
\usepackage{amsmath,amssymb,amsfonts}
\usepackage[linesnumbered,ruled,vlined]{algorithm2e}
\usepackage{graphicx}
\usepackage{textcomp}
\usepackage{xcolor}
\usepackage{listings}
\usepackage{booktabs}
\usepackage{arydshln}
\usepackage{svg}
\hypersetup{colorlinks=true,linkcolor=blue,urlcolor=blue}

\def\BibTeX{{\rm B\kern-.05em{\sc i\kern-.025em b}\kern-.08em
    T\kern-.1667em\lower.7ex\hbox{E}\kern-.125emX}}
\begin{document}

\makeatletter 
\newcommand{\linebreakand}{%
  \end{@IEEEauthorhalign}
  \hfill\mbox{}\par
  \mbox{}\hfill\begin{@IEEEauthorhalign}
}
\makeatother 

\title{LLM-MedQA: Enhancing Medical Question Answering through Case Studies  in \\ Large Language Models \\
}


\author{
\IEEEauthorblockN{Hang Yang}
\IEEEauthorblockA{
CUIT\\
3230608002@stu.cuit.edu.cn}
\and
\IEEEauthorblockN{Hao Chen}
\IEEEauthorblockA{
CUIT\\
haochen@cuit.edu.cn}
\and
\IEEEauthorblockN{Hui Guo}
\IEEEauthorblockA{
University at Buffalo\\
hguo8@buffalo.edu}
\and
\IEEEauthorblockN{Yineng Chen}
\IEEEauthorblockA{
University at Albany\\
ychen77@albany.edu}
\and
\linebreakand
\IEEEauthorblockN{Ching-Sheng Lin}
\IEEEauthorblockA{
Tunghai  University\\
cslin612@thu.edu.tw}
\and
\IEEEauthorblockN{Shu Hu}
\IEEEauthorblockA{
Purdue University\\
hu968@purdue.edu}
\and
\IEEEauthorblockN{Jinrong Hu}
\IEEEauthorblockA{CUIT\\
hjr@cuit.edu.cn}
\and
\IEEEauthorblockN{Xi Wu}
\IEEEauthorblockA{CUIT\\
wuxi@cuit.edu.cn}
\and
\IEEEauthorblockN{Xin Wang}
\IEEEauthorblockA{
University at Albany\\
xwang56@albany.edu}
}

\maketitle

\begin{abstract}
Accurate and efficient question-answering systems are essential for delivering high-quality patient care in the medical field. While Large Language Models (LLMs) have made remarkable strides across various domains, they continue to face significant challenges in medical question answering, particularly in understanding domain-specific terminologies and performing complex reasoning. These limitations undermine their effectiveness in critical medical applications. To address these issues, we propose a novel approach incorporating similar case generation within a multi-agent medical question-answering (MedQA) system. Specifically, we leverage the Llama3.1:70B model, a state-of-the-art LLM, in a multi-agent architecture to enhance performance of classification on the MedQA dataset using zero-shot learning. Our method capitalizes on the model's inherent medical knowledge and reasoning capabilities, eliminating the need for additional training data. Experimental results show substantial gains over existing benchmark models, with improvements of 7\% in both accuracy and F1-score across various medical QA tasks. Furthermore, we examine the model’s interpretability and reliability in addressing complex medical queries. This research not only offers a robust solution for medical question answering but also establishes a foundation for broader applications of LLMs in the medical domain.
\end{abstract}


\section{Introduction}
\begin{figure*}[tb]
\centerline{\includegraphics[scale=0.9,trim= 0 40 0 0,clip]{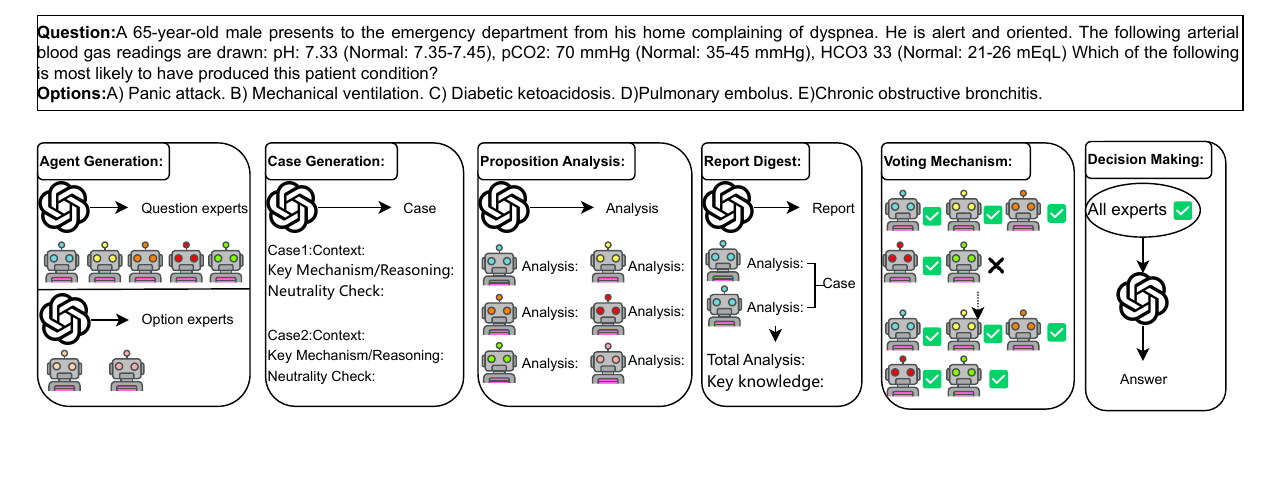}}
\caption{
Illustration of our proposed multi-agent architecture diagram, given a medical problem as an input to the larger model, which is divided into six phases: (1) Agent Generation, (2) Case Generation, (3) Proposition Analysis, (4) Report Digest, (5) Voting Mechanism, and (6) Decision Making.
}
\label{fig-str}
\end{figure*}
The advent of Large Language Models (LLMs)  \cite{dubey2024llama, achiam2023gpt} has revolutionized the field of natural language processing, offering unprecedented capabilities in understanding and generating human-like text across a multitude of domains. However, the medical domain poses unique challenges due to its specialized terminology, complex reasoning requirements, and the critical importance of accuracy in patient care. Medical Question Answering (QA) systems \cite{reichenpfader2024large, yagnik2024medlm} which are designed to provide accurate and reliable information in response to medical queries, must navigate this complexity while ensuring the safety and efficacy of the information delivered. Despite significant advancements, existing LLMs often struggle with the nuances of medical language and the need for precise reasoning, which is essential for high-quality medical QA.

In this study, we introduce a multi-agent framework, as shown in Fig.~\ref{fig-str}, to tackle the QA task on the medical domain. This framework leverages the Llama3.1:70B model, the off-the-shelf LLM model with 70 billion parameters, to enhance the performance of medical QA systems on the MedQA dataset \cite{app11146421}. Our multi-agent approach incorporates specialized agents to handle the inherent complexity of medical QA \cite{zhou2023survey}. Each query in the system is assigned a series of experts, including question-specific analysis, option analysis, and case generation. A key innovation of our multi-agent system is the integration of a case generation module. This module autonomously generates supportive clinical cases tailored to the given question and selected options. Its purpose is to produce plausible and contextually accurate cases that substantiate the correct option for a specific medical problem. These cases, which are integrated into the reporting module, address a critical gap in existing systems by providing transparent and contextually rich explanations. 

Furthermore, the system leverages the Llama3.1:70B model's vast capacity for zero-shot learning, enabling it to reason through complex and specialized medical queries without requiring additional training examples. This capability is especially valuable for the MedQA dataset, where annotated data is scarce and costly to generate, allowing the system to adapt to diverse scenarios with minimal data preparation.

Our primary research objective is to demonstrate the effectiveness of the Llama3.1:70B model, combined with a multi-agent framework \cite{sun2024llm,guo2024large}, in addressing the challenges of medical QA. Specifically, we aim to illustrate how the architectural advantages of both the model and the multi-agent system contribute to improved accuracy, reliability, and interpretability in handling medical queries. Through a series of experiments, we evaluate the model's performance on the MedQA dataset and compare it with other state-of-the-art models. The contributions of the paper are summarized as follows.

\begin{itemize}
    \item  We introduce a novel concept of case studies in the context of a multi-agent medical QA system. The case generation module autonomously generates supportive clinical cases based on the problem and selected options. This approach enhances system interpretability by offering contextually rich and human-readable justifications.
    \item  We present a detailed process for generating supportive clinical cases, which involves extracting key clinical features, such as symptoms and diagnostic findings, from the problem and selected options. Each case consists of three components: Context, Key Mechanism/Reasoning, and Neutrality Check. This ensures that the generated cases are realistic, neutral, and aligned with the correct option, thus supporting the final diagnosis and enhancing interpretability.
    \item  We conduct experiments on the MedQA dataset to evaluate the performance of our multi-agent system. The results show that integrating the case generation module significantly improves the system’s accuracy and interpretability, offering more contextually rich explanations for medical problems.
\end{itemize}

The remainder of this paper is organized as follows: Section II reviews the related work in the field of medical QA and LLMs for medical problem solving and multi-agent systems. Section III describes our methodology, including the model architecture, multi-agent framework. Section IV presents our experimental process and results. Finally, Section V concludes the paper and suggests directions for future research.

\section{Related Work}
\subsection{LLMs for Medical Problem Solving}\label{A}
In recent years, large language models have brought transformative changes to the medical field \cite{karabacak2023embracing}, reshaping key areas such as diagnostics, treatment planning, and communication between healthcare professionals and patients \cite{haque2023future}. 
By assisting physicians in symptom analysis and disease diagnosis \cite{wang2024artificial}, LLMs enhance the accuracy and efficiency of medical assessments \cite{clusmann2023future,prabhod2023integrating}. These models are also capable of supporting clinical decision-making by providing evidence-based recommendations tailored to individual patients \cite{nazi2024large}. 
Additionally, LLMs play a pivotal role in synthesizing and summarizing complex medical information, making it more accessible to both medical professionals and patients. 
Their ability to convey medical advice in a clear and understandable manner significantly improves doctor-patient communication \cite{nerella2023transformers,geantua2024potential}. Moreover, LLMs facilitate the electronic documentation of patient records, streamlining administrative tasks and improving workflow efficiency\cite{menzies2024ai}. Collectively, these advancements highlight the indispensable role of LLMs in optimizing healthcare delivery and outcomes \cite{prabhod2024role}.

Traditionally, enhancing the performance of LLMs in specialized medical tasks has relied heavily on fine-tuning with domain-specific datasets\cite{christophe2024med42}. This process involves curating high-quality medical data and adapting pre-trained models through transfer learning, allowing the models to perform more effectively in new and complex tasks \cite{kim2023predicting, yu2022transfer}. While fine-tuning has proven effective in refining the capabilities of a single model, it requires substantial computational resources and extensive retraining \cite{han2021pre,raffel2020exploring}. However, novel approaches are emerging that bypass the need for additional training, offering a more efficient and cost-effective alternative \cite{brown2020language,luo2024zero}. These approaches enable healthcare providers to benefit from advanced model applications without the need for extensive customization, thus making these tools more accessible across different medical environments\cite{erion2022cost}.

In contrast to the traditional single-model fine-tuning approach, the multi-agent system offers a more robust framework for medical decision-making. By enabling multiple agents to collaborate, exchange information, and analyze clinical cases from diverse perspectives, multi-agent systems enhance the accuracy and reliability of medical decisions\cite{tang2023medagents}. This collaborative approach harnesses the collective intelligence of various agents, resulting in more comprehensive and well-informed clinical outcomes \cite{yue2024ct}.

\subsection{Multi-Agent Systems for Medical Decision-Making}
A multi-agent system is a system that coordinates and collaborates with multiple autonomous agents to accomplish tasks together\cite{liu2024multi}. 
These intelligent agents can solve complex problems more efficiently than a single agent through information sharing, role allocation, and feedback mechanisms\cite{guo2024large,wang2023deep}. 
In such a system, each agent can play different roles and analyze and handle problems from different perspectives.

Multi-agent systems have demonstrated their superior problem-solving capabilities across various domains. For instance, in the financial sector, multiple agents monitor real-time market dynamics and make investment decisions based on a variety of market signals\cite{han2024enhancing}. 
In logistics, intelligent agents collaborate to coordinate transportation and distribution processes, optimizing supply chain management\cite{khayyat2016intelligent}. 
These examples showcase the strength of multi-agent systems in handling dynamic environments and complex decision-making tasks\cite{kimmdagents}.

In the context of medical decision-making, several factors such as a patient's medical history, physical examination data, multi-modality data \cite{wang2024neural}, and expertise from multiple medical specialties must be integrated\cite{isern2016systematic}. 
Traditional decision-making systems often struggle to manage this complexity \cite{wang2024u}, but multi-agent systems address these challenges by leveraging role allocation and feedback mechanisms across different dimensions. 
This approach significantly enhances the precision and reliability of medical decisions\cite{bhanu2022applications}.

\section{Methodology}

\begin{figure*}[tb]
\centerline{\includegraphics[scale=0.58, trim= 30 100 40 0,clip ]{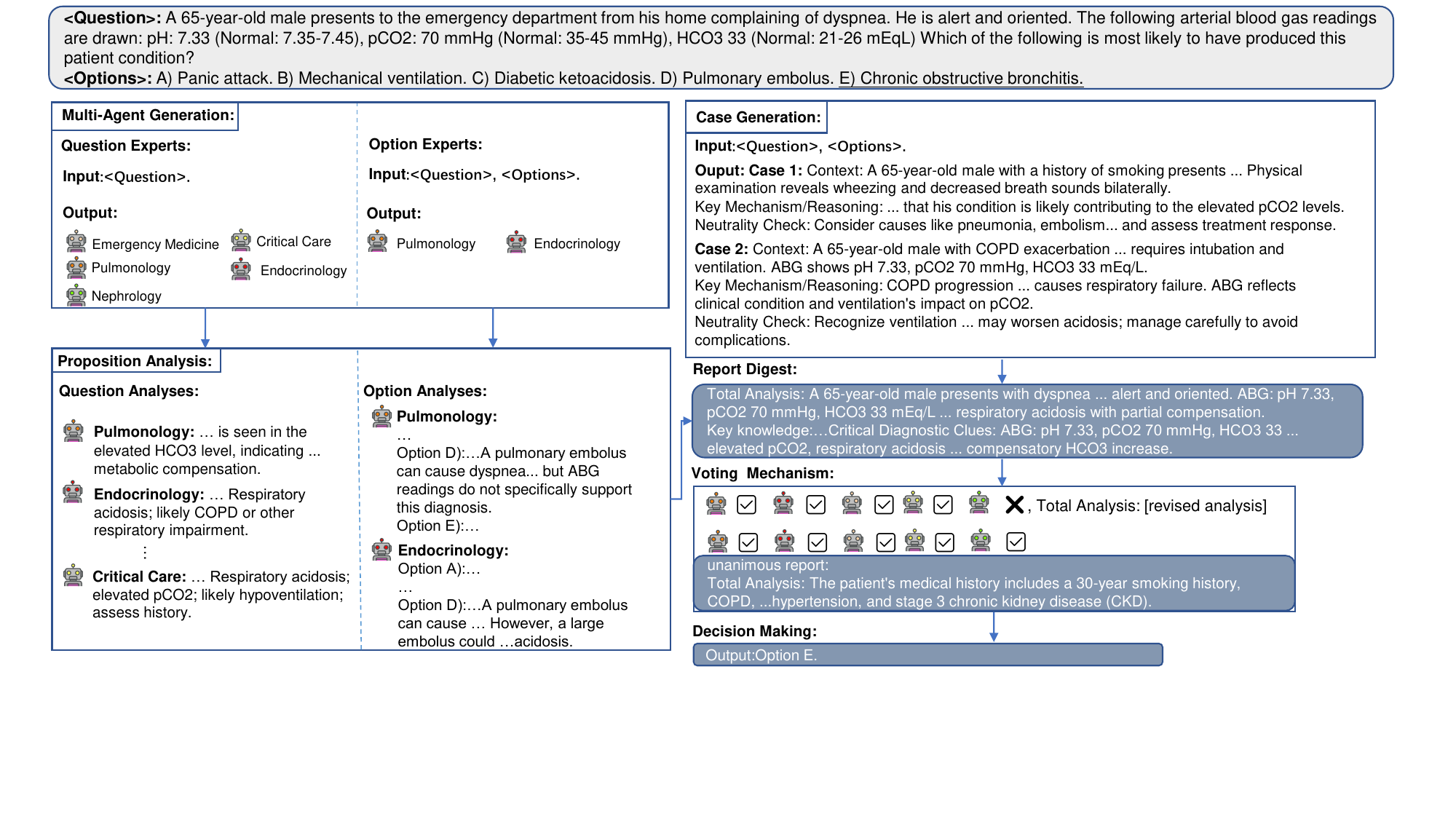}}
\caption{
Diagram of the Proposed Multi-Agent System Framework for Medical Question Answering
}
\label{fig-str1}
\end{figure*}
In this section, we propose a specific multi-agent model framework (Fig. \ref{fig-str}) to tackle the task of Medical Question Answering. The overall model consists of six components: (1) \textit{Multi-agent generation}, which includes the creation of question experts and option experts. (2) \textit{Proposition analysis}, this step involves a detailed analysis of the problem and the available options. (3) \textit{Case generation}, which generates relevant cases based on the input questions and provided options. (4) \textit{Report digest}, a report is generated by synthesizing insights from problem analysis, option analysis, and case generation. (5) \textit{Voting mechanism}, where experts vote on the generated report, revising it as necessary if disagreements arise. This process continues iteratively until consensus is reached; and (6) \textit{Decision making}, where the final report is used as the basis for selecting the correct answer. Apart from above components we described, we finally provide an overview of the algorithm used to facilitate expert voting and decision-making, and model selection explains why we chose to use the LLama 3.1:70B model.
\subsection{Multi-Agent Generation}\label{AA}
In the context of clinical medical problems, given a problem \( q \) and a set of options \( \text{op} = \{ o_1, o_2, \ldots, o_k \} \), where \( k \) denotes the total number of available options, the goal of this process is to assemble a team of experts. These include question experts, specialized in clinical problem analysis, \( \text{questionExperts} = \{ \text{qe}_1, \text{qe}_2, \dots, \text{qe}_m \} \) , as well as option experts specialized in analyzing the options, \( \text{optionExperts} = \{ \text{oe}_1, \text{oe}_2 ,\ldots , \text{oe}_n \} \), where \( m \) and \( n \) represent the respective numbers of question and option domain experts. Specifically, we assign a prompt to the model and provide instructions to guide it in generating the corresponding domain experts based on the input problems and options:
\begingroup
\setlength{\abovedisplayskip}{5pt}  
\setlength{\belowdisplayskip}{5pt}  
{
\begin{equation}
    questionExperts = \text{GenerateExpert}(q, prompt_{qe})
\end{equation}
}
\endgroup
\begingroup
\setlength{\abovedisplayskip}{5pt}  
\setlength{\belowdisplayskip}{5pt}  
{
\begin{equation}
    optionExperts=\text{GenerateExpert}(q,op, prompt_{oe})
\end{equation}
}
\endgroup

Two prompts in equations are represented for generating the question experts and option experts respectively. They guide the model’s behavior during the expert generation process, ensuring LLM performs the appropriate categorization tasks based on the given problem and options. To be specific, 
\textbf{$\text{prompt}_{\text{qe}}$} uses the format: \textit{Description \textless $prompt_{qe}$: "You need to classify the following question into one subfield of medicine based on the given medical scenario: '''\{question\}'''. Consider relevant diagnoses and related fields. Provide the classification in the format '''\{question\_domain\_format\}''', keeping your response concise and under \{max\_words\} words."
\textgreater}, \textbf{$\text{prompt}_{\text{oe}}$} uses the format: \textit{Description \textless $prompt_{oe}$: Classify the following options: '''\{options\}''', based on the medical scenario: '''\{question\}'''. Output them in the format '''\{options\_domain\_format\}'''."
\textgreater }

\textbf{Question Domain Experts:} These experts specialize in
clinical knowledge related to specific medical issues. They
analyze symptoms, diagnoses, and treatment options, provid-
ing critical insights for decision-making. This group
includes specialists from fields such as infectious diseases,
gynecology, and hematology, and is crucial in identifying
features requiring immediate attention, ensuring patient safety
and care.

\textbf{Option Domain Experts:} These experts analyze the clinical options available for a specific medical issue. Their primary role is to assess the relevance and correctness of each option, considering the nuances between them. By leveraging their extensive clinical experience, they help identify misleading options and provide critical insights that guide the team in selecting the most appropriate treatment pathways.

\subsection{Proposition Analysis}
\textbf{Question Analyses:} After consulting with the experts from relevant fields regarding the problem, we asked them to provide their individual analyses, which are then used to inform further reasoning. For each question \( q \) and corresponding question expert $ qe_i \in questionExperts$, we employ a LLM to act as a domain-specific expert. Guided by the prompt $prompt_{qa}$, the LLM generates an analysis, represented by the following equation:
\begingroup
\setlength{\abovedisplayskip}{5pt}  
\setlength{\belowdisplayskip}{5pt}  
{
\begin{equation}
 qA_i = \text{AnalyzeQuestion}(q, qe_i,prompt_{qa})
\end{equation}
}

The \textbf{$\text{prompt}_{\text{qa}}$} directs the LLM to: (1) Identify the key components of the question, such as symptoms, potential diagnoses, and treatment options; (2) Highlight any critical or urgent features that require immediate attention; (3) Offer a structured analysis, outlining the logical connections between symptoms, diagnosis, and recommended next steps.

\textbf{Option Analyses:} Once the question analysis is complete, we proceed to evaluate the options provided. This process involves examining the relationships among the options as well as their relevance to the question. For each option analysis \(oA_i\), the LLM is supplied with the question \(q\), the option \(op\), a specific option domain expert $oe_i \in optionExperts$, and the previously generated question analysis 
\( qA_i \) (produced by the question domain expert \( qe_i \)). The LLM generates the option analysis based on this input as follows:
\begingroup
\setlength{\abovedisplayskip}{5pt}  
\setlength{\belowdisplayskip}{5pt}  
{
\begin{equation}
 oA_i = \text{AnalyzeOption}(q, op, oe_i, qA_i, prompt_{oa})
\end{equation}
}
\endgroup

The \textbf{$\text{prompt}_{\text{oa}}$} directs the LLM to: (1) Each option needs to be analyzed independently to assess its relevance to the patient's clinical situation and the available evidence. (2) Analyze the reasonableness of the options to determine if they are the most appropriate next step or should be excluded. (3) The analysis should to consider both supporting and opposing evidence to ensure objectivity, independent of the analysis part of the question.

In this process, option domain experts analyze each option to assess their relevance and correctness in relation to the question. Drawing on their medical expertise, they evaluate the validity of the options, identify potentially misleading ones, and provide detailed reasoning on whether each option should be accepted or excluded.

\subsection{Case Generation}
A key component of our LLM-MQA system is the case generation. We introduce it to serve as supportive evidence that aids in the final diagnosis and enhances the interpretability of the overall system. The generated cases are not standalone outputs but work synergistically with the analyses from the problem and option experts. They offer context-rich, interpretable explanations that justify the recommended diagnosis or treatment and are seamlessly integrated into the final report. In addition, the entire system provides not only the correct answer but also a clear and well-supported reasoning process, improving both accuracy and interpretability.

During the case generation phase, the LLM to autonomously creates clinical cases that align with a plausible and correct option based on the dataset. The LLM begins by identifying key clinical features, such as symptoms, examination findings, laboratory results, and other diagnostic factors. Using these elements, it generates one or two concise, realistic clinical cases. These cases are intended for use in the report generation phase, where they provide additional context to support final decision-making. Each generated case consists of  the following components:

\textbf{Context:} Provides a detailed clinical scenario, highlighting key symptoms, medical history, and diagnostic findings.

\textbf{Key Mechanism/Reasoning:} Justifies the selected option by explaining how the clinical findings support the correct diagnosis or treatment, emphasizing the alignment between the case and the chosen outcome.

\textbf{Neutrality Check:} Maintains objectivity by avoiding exaggerated claims about the selected option, while briefly acknowledging relevant alternatives when appropriate.

The LLM follows a structured \textbf{$\text{prompt}_{\text{exa}}$} to guide the generation of these cases: Analyze the question and options to identify the most plausible correct option. Generate 1-2 concise cases: Highlight the clinical reasoning behind the selected option. Provide relevant clinical context, focusing on symptoms, diagnostic findings, or treatments. Present a balanced view by avoiding overemphasis on the correct option while acknowledging alternatives where appropriate. The generation process is represented by the following equation:
\begingroup
\setlength{\abovedisplayskip}{5pt}  
\setlength{\belowdisplayskip}{5pt}  
{
\begin{equation}
 exa_i = \text{GenerateCase}(q,op,prompt_{exa})
\end{equation}
}
\endgroup

\subsection{Report Digest}
In the Report Generation phase, the LLM plays a crucial role as a "synthesizer",  integrates insights derived from the different analysis modules—Question Analysis (QA), Option Analysis (OA), and Case Generation (CG). This phase is designed to create a coherent, well-supported report that is not only accurate but also interpretable. The generated clinical cases, produced by the Case Generation module, are particularly important because they provide contextually rich, clinical justifications for the selected options. These cases add depth to the final report, enhancing its transparency and interpretability.

In this process, the LLM first extracts the key information from each analysis module and identifies areas of agreement and disagreement among the experts. It then synthesizes these insights and generates a comprehensive report offering a nuanced and complete view of the problem. The LLM carefully balances the clinical data from the analyses with the generated cases to form a cohesive and informative report. The cases contribute significantly to this synthesis by grounding the theoretical analyses in real-world clinical contexts. In generating the report, LLM follows a structured \textbf{$\text{prompt}_{\text{Rp}}$} which requires extracting key information from the problem analysis, option analysis, and case study analysis, and generating two core sections of the report based on this information:

\textbf{Key Knowledge:} In this section, the most important diagnostic clues, clinical context, and reasoning are extracted from all three modules: Question Analysis, Option Analysis, and Case Generation. The Case Generation module plays a pivotal role here, as it provides detailed clinical scenarios that are aligned with the correct options, offering concrete examples that illustrate the reasoning behind the conclusions. This section ensures that all analyses are accurately represented and highlights the most relevant information to support the decision-making process.

\begin{algorithm}[t]
\label{alg:1}
\caption{LLM-MedQA Report Process}
\KwIn{Expert group $E = \{e_1, \dots, e_m\}$, Initial report $R_0$, Feedback model $M$, Maximum iterations $k$, Interaction prompts $\{p_{vote}, p_{modify}, p_{revise}\}$}
\KwOut{Final report $R_f$}

\SetKw{KwInitialize}{Initialize: $flag\_feedback\_required \gets \text{True}$, $iteration\_count \gets 0$, $c\_draft \gets R_0$, $suggestions \gets \emptyset$}
\SetKw{KwIncrement}{Increment: $iteration\_count \gets iteration\_count + 1$}
\SetKw{KwResetFlag}{Reset: $flag\_feedback\_required \gets \text{False}$}

\SetKw{KwGetVote}{\text{Collect feedback:} \\
\text{$vote_i \gets M(c\_draft, e_i, p_{vote})$}}
\SetKw{KwGetSuggestions}{\text{ Generate suggestion:} \text{$suggestion_i \gets M(current\_draft, e_i, p_{modify})$}}
\SetKw{KwAccumulate}{\text{Accumulate suggestions:} \text{$suggestions \gets suggestions + suggestion_i$}}
\SetKw{KwSetFlag}{\text{Mark feedback required:} \text{$flag\_feedback\_required \gets \text{True}$}}

\SetKw{KwRevise}{\text{Revise report:}\\
\text{$cc\_draft \gets M(c\_draft, suggestions, p_{revise})$}}
\SetKw{KwReturnFinal}{\text{Return final report:} \text{$R_f \gets c\_draft$}}

\KwInitialize{}

\While{$flag\_feedback\_required$ and $iteration\_count < k$}{
    \KwIncrement{}
    \KwResetFlag{}
    
    \For{each expert $e_i$ in $E$}{
        \KwGetVote{}
        
        \If{$vote_i = \text{disagree}$}{
            \KwGetSuggestions{}\\
            \KwAccumulate{}\\
            \KwSetFlag{}
        }
    }
    
    \If{$flag\_feedback\_required$}{
        \KwRevise{}
    }
}
\KwReturnFinal{}
\end{algorithm}

\textbf{Total Analysis:} This section synthesizes the entire clinical scenario by incorporating clinical features from the Case Generation module. It evaluates each option by considering both supporting and refuting evidence and ranks them based on their clinical relevance. The generated cases ensure that the evaluation is grounded in realistic clinical situations, enabling a direct comparison of options within the context of the problem. The LLM then provides a ranked recommendation with clear justification, grounded in both the analyses and the generated cases. The process can be represented by the following equation:
\begingroup
\setlength{\abovedisplayskip}{5pt}  
\setlength{\belowdisplayskip}{5pt}  
{
\begin{equation}
 Repo = \text{GenerateReport}(qA, oA, exa, prompt_{Rp}) \end{equation}
}
\endgroup

The Case Generation module plays a critical role in this process. By generating clinical cases that reflect the correct options and relevant alternatives, it ensures that the report provides not only  theoretical analysis but also a realistic clinical perspective. This adds depth to the final report, making it more interpretable and transparent, while providing healthcare professionals with clear, evidence-backed explanations that aid in decision making.

Thus, the Report Generation module, enhanced by the Case Generation module, creates a comprehensive, coherent, and interpretable report that offers objective recommendations for the final diagnosis or treatment option. The cases contribute to the overall narrative by illustrating how the clinical findings align with the selected option, clarifying the reasoning behind the recommendations.
\subsection{Voting Mechanism}

After generating the report, we implement a voting decision-making mechanism with "Yes" and "No" as the voting options. If the experts find the report unreasonable, they will cast a negative vote ("No") and provide revision suggestions to address the identified issues. Conversely, if the experts unanimously agree that the report is reasonable ("Yes"), we proceed to the next stage of selecting the correct answer.

To ensure the quality of the report, the comprehensive report (\textit{Repo}) is submitted to all participating experts, including both question domain experts and option domain experts. The voting process involves each expert casting a vote of either "Approve" ("Yes") or "Reject" ("No").

If all experts vote "Approve" ("Yes"), the report is considered reasonable, and we proceed to the next stage. If any expert votes "Reject" ("No"), their feedback and revision suggestions are collected. The report is then revised and resubmitted for re-voting until unanimous approval is achieved.

To facilitate the revision process and ensure consistency in the incorporation of expert feedback, we use a structured prompt called $prompt_{mod}$. This $prompt_{mod}$ guides the model in generating the modified report based on the feedback provided by the experts. Specifically, \textbf{$\text{prompt}_{\text{mod}}$} is designed to:

Integration of Expert Feedback: It takes the original report and the feedback provided by each expert in different domains (e.g., question domain, option domain) and incorporates them into the revised version.
Ensure consistent format: It directs the model to follow a specific format when making modifications, ensuring that the final report maintains its structural integrity and aligns with the expectations of the experts.
The revision process, guided by promptmod, can be represented by the following equation:
\begingroup
\setlength{\abovedisplayskip}{5pt}  
\setlength{\belowdisplayskip}{5pt}  
{
\begin{equation} Repo_i = \text{ModifyReport}(Repo_{i-1}, Mod_i, prompt_{mod}) 
\end{equation}
}
\endgroup

Where \( \textit{Repo}_{i-1} \) is the previously revised report, \( \textit{Mod}_i \) is the modification based on expert feedback, and \( \textit{prompt}_{mod} \) is the prompt used to guide the generation of the modified report.

This process ensures that the final report, serving as the foundation for selecting the correct answer, reflects the collective consensus of the experts and maintains the highest quality.

\subsection{Decision Making}
In the final step, we require the LLM to act as the medical decision maker, deriving the final answer to the clinical question \( q \) based on the unanimous report \( Repo_f \). The decision-making process can be represented by the following equation:
{
\begin{equation}
     output = \text{MakeDecision}(q, op, Repo_f, prompt_{dm})
\end{equation}
}

The \textbf{$\text{prompt}_{\text{dm}}$} directs the model to: Review the synthesized report and identify the most supported option. If no option is clearly confirmed, evaluate each option based on its alignment with the findings in the report, the patient’s clinical context, and general medical reasoning. In cases where multiple options are plausible, eliminate less supported options and prioritize the most consistent one.

The entire process of the Voting Mechanism is summarized into an Algorithm \ref{alg:1} to provide a clearer understanding of the LLM's operational process.

\section{Experiments}
\subsection{Dataset and Evaluation Metric}\label{B}
We conducted experiments on the publicly available \textbf{MedQA}\cite{app11146421} dataset, which is specifically designed for questions and answers in the medical field. The dataset consists of multiple-choice medical questions, as detailed in Fig.~\ref{fig-str1}. Each instance includes a clinical query, a set of five answer options, and a correct answer for validation purposes.
The MedQA dataset presents unique challenges due to the specialized nature of medical knowledge and the complexity of reasoning required to derive the correct answers. It provides a useful testbed for evaluating medical question answering systems, particularly in the context of leveraging large language models.

Considering the ultimate goal of model is to identify the best option from the multiple choices. To comprehensively evaluate the performance of our proposed system and compare it against other baselines, we adopt four widely used evaluation metrics in multi-class classification task:
\begin{itemize}
    \item \textbf{Accuracy}: This measures the overall proportion of correct predictions. For multi-class classification, it is computed by summing the correct predictions (true positives) across all classes and dividing by the total number of samples.
    
    \item \textbf{Macro Precision} is the average of precision scores across all classes, without considering the class distribution. The formula is $\frac{1}{C}\sum_{C}^{i=1}Precision_{i}$, where $C$ is the number of classes, and $Precision_{i}$ is the $precision$ for the $i-th$ class.
    
    \item \textbf{Macro Recall} is the average of recall scores across all classes, without considering the class distribution. The formula is $\frac{1}{C}\sum_{C}^{i=1}Recall_{i}$, where $C$ is the number of classes, and $Recall_{i}$ is the $recall$ for the $i-th$ class.
    
    \item \textbf{Macro F1-Score} is the average of F1-scores across all classes, without considering the class distribution. The formula is $\frac{1}{C}\sum_{C}^{i=1}F1_{i}$, where $C$ is the number of classes and $F1_{i}$ is the $F1$ for the $i-th$ class calculated by the equation:
    \begin{equation}
        2 * \frac{Precision_{i}*Recall_{i}}{Precision_{i}+Recall_{i}}
    \end{equation}
\end{itemize}

\subsection{Experiments Settings}
\textbf{Configuration Settings} In our experiments, we used the off-the-shelf LLM that is developed by Meta's FAIR team. It is an open-source model and easy to deploy within Llama framework. All experiments were conducted in a \textbf{zero-shot} setting. We randomly selected 300 samples from the dataset three times, and the final experimental results are reported as the average of these three runs. The four key inference parameters—temperature, frequency penalty, and presence penalty—are all set to 0, and top\_p is set to 1. The inference time for each example in our method is approximately one and a half minutes.

In addition, we have chosen the Llama3.1:70B model due to its open-source nature, relatively low computational requirements, and cost-effectiveness compared to other models of similar performance. This choice ensures that the model remains accessible for research and application while providing high performance, making it particularly suitable for real-world healthcare QA scenarios. 

Although results from the few-shot setting are reported in the table, these settings were used as a baseline for comparison experiments. We selected few-shot as the baseline to evaluate the model's performance with a small number of examples, allowing us to compare the performance against the zero-shot setting and analyze the extent of improvement.

\subsection{Baselines} We used the vLLM \cite{kwon2023efficient} technique to easily access the model with the following baseline: \textbf{(1) Direct Inference} involves providing the question and its possible answer options directly as input to the large language model. The model then generates a response based on its internal knowledge, without additional reasoning or thought processes. This method is straightforward and computationally efficient, making it ideal for scenarios where questions are simple and well-defined. It can be applied in both zero-shot and few-shot settings. In the few-shot setting, providing a small number of example questions and answers helps the model generalize better, potentially leading to more accurate results compared to the zero-shot case. However, even in the few-shot setting, Direct Inference remains limited by the model’s reliance on pre-trained knowledge without deeper reasoning steps. \textbf{(2) CoT}, the Chain of Thought (CoT) method enhances the reasoning ability of LLMs by encouraging them to work through problems step by step, simulating a more human-like process of thought. Instead of directly outputting an answer, the model is prompted to generate intermediate reasoning steps that lead to the final answer. This method is particularly effective for complex questions that require multi-step analysis. It can be used in both zero-shot and few-shot settings. In the few-shot case, providing examples of reasoning steps helps the model better understand how to approach similar problems, improving its performance compared to the zero-shot case. However, CoT can be resource-intensive and may introduce additional complexity in the output.  \textbf{(3) CoT+SC}, the Chain of Thought with Self-Consistency (CoT+SC) method builds upon CoT by introducing the self-consistency technique, which involves generating multiple reasoning paths and selecting the most consistent answer. This method leverages multiple reasoning attempts to improve the robustness and reliability of the model's final answer. It can be applied in both zero-shot and few-shot settings. In the few-shot setting, providing multiple examples of reasoning steps and answers helps the model produce more consistent and accurate outputs. However, this approach is more computationally expensive as it requires the model to generate and compare multiple outputs.

\begin{table}[t]
    \centering
    \caption{Comparison of Our Methods at Baseline} 
    \scalebox{0.92}{
    \begin{tabular}{lcccc}
        \toprule
        \textbf{Method} & \textbf{Accuracy} & \textbf{\makecell{Macro\\Precision}} & \textbf{\makecell{Macro\\Recall}} & \textbf{\makecell{Macro\\F1-Score}} \\
        \midrule
        \multicolumn{5}{l}{\textit{*few-shot setting}} \\
        Direct Inference\cite{brown2020language} & 0.717 & 0.717 & 0.715 & 0.715 \\
        CoT \cite{wei2022chain} & 0.710 & 0.708 & 0.709 & 0.708 \\
        CoT+SC \cite{wang2022self} & 0.727 & 0.725 & 0.724 & 0.724 \\
        \midrule
        \multicolumn{5}{l}{\textit{*zero-shot setting}} \\
        Direct Inference \cite{brown2020language} & 0.714 & 0.715 & 0.714 & 0.713 \\
        CoT \cite{wei2022chain} & 0.698 & 0.697 & 0.697 & 0.697 \\
        CoT+SC \cite{wang2022self} & 0.719 & 0.719 & 0.719 & 0.718 \\
        \textbf{Ours} & \textbf{0.772} & \textbf{0.771} & \textbf{0.772} & \textbf{0.771} \\
        \bottomrule
    \end{tabular}
    }
    \label{tab:comparison_baseline}
    \vspace{1mm} 
    \makebox[\textwidth][c]{
       \parbox{0.95\textwidth}{\footnotesize Table 1: SC denotes the self-consistency prompting method.\\ Results in \textbf{bold} indicate optimal performance.}%
}
\end{table}

\subsection{Result and Analysis}
Tab.~\ref{tab:comparison_baseline} shows the overall results of comparison. It can be observed that our method achieves the best across all metrics, indicating that our multi-agent architecture offers advantages in  real-world medical scenarios. In the few-shot setting, the Direct Inference method achieves an accuracy of 0.717, with Macro Precision, Recall, and F1-Score all at 0.71. Incorporating CoT results in a slight performance decrease, with accuracy dropping to 0.710 and macro metrics around 0.708. Adding SC to CoT further enhances performance, achieving the best results in the few-shot setting with an accuracy of 0.727 and macro metrics around 0.724. In the zero-shot setting, the baseline method performs relatively well with an accuracy of 0.714 and macro metrics of 0.715, while the CoT method alone demonstrates significant performance degradation, achieving only 0.698 in accuracy and 0.697 in macro metrics. The addition of SC to CoT shows notable improvement, achieving an accuracy of 0.719 and macro metrics of 0.718. The proposed "Ours" method outperforms all other approaches in both zero-shot and few-shot settings, achieving the highest scores across all metrics, with an accuracy of 0.772 and macro metrics of 0.771. These results highlight the robustness and effectiveness of the proposed method, particularly in the challenging zero-shot scenario. The multi-agent architecture plays a key role in addressing complex decision-making problems. It not only integrates different expert opinions more effectively but also handles uncertainties common in healthcare scenarios, thereby improving the comprehensiveness and reliability of decision-making.

\subsection{Ablation Study}
We conducted an ablation study to assess the impact of different LLM scales on our model's performance. Specifically, we deployed our multi-agent system using two LLM sizes—8B and 70B—with and without the case generation process. The results, summarized in Tab.~\ref{tab:ablation}, reveal that the larger model significantly outperforms the smaller one across all metrics, with performance increasing from approximately 55\% to over 70\%. Based on these findings, we opted to use the 70B model for our overall analysis. Furthermore, the inclusion of the case generation module was shown to enhance the model's performance, likely by providing richer contextual information and improving the system’s ability to classify answers accurately. These results highlight the module’s role in bolstering the selection process and overall system effectiveness.
\begin{table}[t]
\centering
\caption{Ablation Study of LLM Scales with/without Case Generation}
\scalebox{0.9}{
\begin{tabular}{lcccc}
\toprule
        \textbf{Method} & \textbf{Accuracy\%} & \textbf{\makecell{Macro\\Precision}} & \textbf{\makecell{Macro\\Recall}} & \textbf{\makecell{Macro\\F1-Score}}  \\ \midrule
Mutli-Agent(8B)& 56.3 & 55.9 & 55.9 & 55.8 \\
Mutli-Agent(70B)& 73.0 & 73.5 & 72.9 & 73.0 \\
\hdashline
+ Case(8B) & 57.3 ($\uparrow$ 1.0) & 56.9($\uparrow$ 1.0) & 56.8($\uparrow$ 0.9) & 56.8($\uparrow$ 1.0) \\
+ Case(70B) & 75.0 ($\uparrow$ 2.0) & 74.9($\uparrow$ 1.4) & 74.7($\uparrow$ 1.8) & 74.8($\uparrow$ 1.8)\\ \bottomrule
\end{tabular}
}
\label{tab:ablation}
 \vspace{1mm} 
    \makebox[\textwidth][c]{\parbox{0.97\textwidth}{\footnotesize Table 2:Ablation study for LLM scales with/without case generation.}}
\end{table}


\section{Conclusion}
In this paper, we introduce a multi-agent framework for medical question answering (MedQA), leveraging specialized domain experts, a case generation module, and a voting mechanism to enhance decision-making. Our approach integrates domain experts with a case generation module that uses clinical data to support the selection of the most plausible answers. The framework employs a joint optimization mechanism, where feedback from domain experts and the case generation module is utilized to refine problem and option analysis tasks, feeding insights back into the large language model. Additionally, a voting mechanism aggregates expert feedback and revisions, improving the quality and reliability of the generated reports.

Comprehensive experiments on the MedQA dataset demonstrate that our approach outperforms existing methods, such as direct inference and Chain of Thought (CoT), across key metrics, including accuracy, precision, recall, and F1-score. Our method achieves nearly 77\% across all metrics, compared to approximately 70\% for other approaches. Furthermore, we validate that employing a large-scale LLM significantly enhances performance, with the case generation step identified as a critical component driving these improvements. This framework enhances the system’s explainability of reasoning and ensures robust decision-making through expert consensus, providing a reliable and effective solution for medical question-answering tasks.

In future work, we aim to further investigate the case generation module to support a wider range of clinical scenarios, incorporating diverse patient profiles and diagnostic complexities. Additionally, we plan to explore the scalability of our multi-agent framework in real-time medical environments, focusing on optimizing model efficiency and response times for clinical practitioners. Our approach offers a solid foundation for building advanced, explainable medical question-answering framework, which is general and can be applied to other complex decision-making tasks.

{
\bibliographystyle{IEEEtran}
\bibliography{IJCNN}
}


\end{document}